\pdfoutput=1

\documentclass[11pt]{article}

\usepackage{acl}

\usepackage{times}
\usepackage{latexsym}
\usepackage{graphicx}

\usepackage[T1]{fontenc}

\usepackage[utf8]{inputenc}

\usepackage{microtype}

\title{Language Diversity: Visible to Humans, Exploitable by Machines}

\author{Gábor Bella \\
  University of Trento\\
  via Sommarive, 5, 38123 Trento, Italy \\
  \texttt{gabor.bella@unitn.it} \\
  \And
  Erdenebileg Byambadorj \\
  University of Trento\\
  via Sommarive, 5, 38123 Trento, Italy \\
  \texttt{e.byambadorj@unitn.it} \\
  \AND
  Yamini Chandrashekar \\
  University of Trento\\
  via Sommarive, 5, 38123 Trento, Italy \\
  \texttt{yamini.chandrashekar@unitn.it} \\
  \And
  Khuyagbaatar Batsuren \\
  National University of Mongolia\\
  NUM 1, 14200, Ulaanbaatar, Mongolia \\
  \texttt{khuyagbaatar@num.edu.mn} \\
  \AND
  Danish Ashgar Cheema \\
  University of Trento\\
  via Sommarive, 5, 38123 Trento, Italy \\
  \texttt{danish.cheema@unitn.it} \\
  \And
  Fausto Giunchiglia \\
  University of Trento\\
  via Sommarive, 5, 38123 Trento, Italy \\
  \texttt{fausto.giunchiglia@unitn.it} \\
}

\def\para#1{\noindent\textbf{#1}}

\begin{document}
\maketitle
\begin{abstract}
The \emph{Universal Knowledge Core} (UKC) is a large multilingual lexical database with a focus on language diversity and covering over a thousand languages. The aim of the database, as well as its tools and data catalogue, is to make the somewhat abstract notion of diversity visually understandable for humans and formally exploitable by machines. The UKC website lets users explore millions of individual words and their meanings, but also phenomena of cross-lingual convergence and divergence, such as shared interlingual meanings, lexicon similarities, cognate clusters, or lexical gaps. The \emph{UKC LiveLanguage Catalogue}, in turn, provides access to the underlying lexical data in a computer-processable form, ready to be reused in cross-lingual applications.
\end{abstract}


\section{Introduction}

A recent challenge in computational linguistics has been the development of efficient multilingual and cross-lingual techniques for language understanding and processing. In terms of solutions, a mainstream, yet often implicit assumption has been that shared meaning unites languages beyond superficial differences in lexicon and grammar: after all, humankind on the whole has been successful in getting ideas across linguistic borders. Hence the recent trend of massively multilingual resources---lexical databases, cross-lingual transfer matrices, pre-trained multilingual language models---exploiting a common meaning-based mapping across linguistic units.

\emph{Linguistic diversity} remains, nevertheless, a key concept insomuch as it refers to deep-running, irreducible, meaning-level differences across languages and underlying cultural concepts. To take real examples from state-of-the-art machine translation, syntactically correct but semantically absurd outputs such as \emph{`my older brother is younger than me'} or \emph{`this raw rice is tasty'} are not rare exceptions but recurrent consequences of diversity: the diverging ways languages express culturally significant concepts such as \emph{brother} or \emph{rice}. While phenomena such as lexical gaps \cite{lehrer1970notes}, culturally diverse terminology, or the varying relevance of the notion of \emph{word} itself across languages are not unfamiliar to us computational linguists, such a general and intuitive understanding is hard to translate into actual `diversity-aware' computational applications, not the least because of the lack of formal datasets that would provide such information.

Lexical typology has described and catalogued many of such phenomena \cite{koptjevskaja2015semantics}. A few online databases also provide contrastive data, sometimes over thousands of languages \cite{wals,clics,asjp}. These databases are rarely used in the NLP community, probably because they are often targeted towards historical linguistics and use phonetic representations of words or are limited to a few hundred core concepts. Yet, our position is that typological data can and should be reused for computational purposes, provided that they are meaningfully integrated with existing resources on contemporary language.

\begin{figure*}[t]
\includegraphics[width=\textwidth]{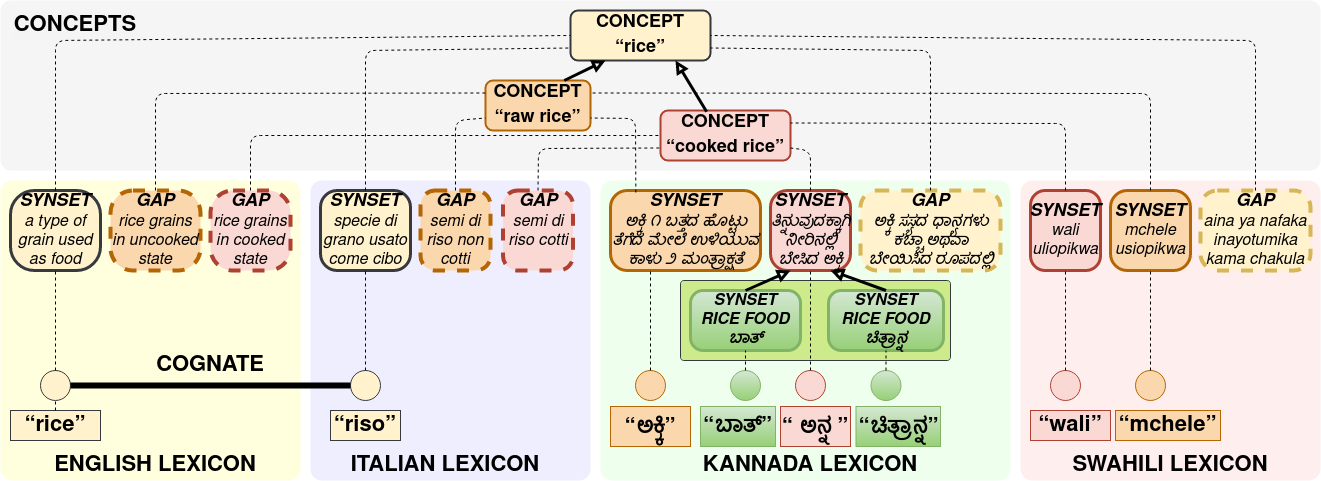}
\caption{\label{fig:ukc_structure}Structural elements in the UKC lexical database for representing cross-lingual unity and diversity.}
\end{figure*}

Computationally-oriented resources that address language diversity or linguistic typology have so far been concentrating on multilingual morphosyntax \cite{ponti2019modeling,batsuren2021morphynet,nivre2016universal}. On diversity in lexical semantics, only a few studies \cite{giunchiglia2017understanding} and sporadic data have been available for specific languages, such as a few hundred lexical gaps in Hebrew \cite{ordan2007hebrew} or in Italian \cite{pianta2002multiwordnet}. Large-scale multilingual lexical databases (MLDB), such as BabelNet \cite{Roberto2012babelnet} or the Open Multilingual Wordnet \cite{bond2013linking}, only model shared meaning and thus do not offer data on lexical diversity.

The \emph{Universal Knowledge Core} (UKC) database and system presented in this paper provides computer-readable cross-lingual lexical data, covering both the common and the diverse among more than a thousand lexicons. The data is being made available for download from the \emph{UKC LiveLanguage} catalogue\footnote{\url{http://www.livelanguage.eu}}, while the \emph{UKC website}\footnote{\url{http://ukc.datascientia.eu}} provides a set of interactive tools that allow both high-level visualisations and an in-depth exploration of diversity data. The rest of the paper provides an overview of the UKC database structure and contents, the online tools, and the data catalogue.\footnote{See \url{http://youtu.be/b90SdCJjtCw} for a video.}

\begin{figure*}[t]
\includegraphics[width=\textwidth]{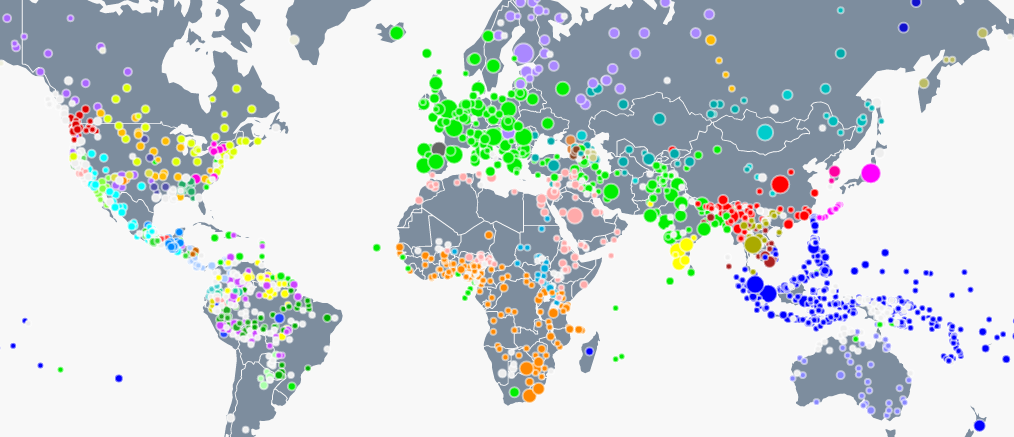}
\caption{\label{fig:ukc_map}Lexicons in the UKC: circle sizes indicate lexicon size and their colour the language family (phylum).}
\vspace{-2mm}
\end{figure*}

\section{A Multilingual Lexical Database on~Language Diversity}

Among existing large-scale MLDBs, those with a published formal, computer-exploitable data model---such as the Open Multilingual WordNet, BabelNet, or EuroWordNet \cite{vossen1997eurowordnet}---concentrate on \emph{language unity}, representing shared meaning through linking together words with the same meaning across languages. The UKC simultaneously enriches existing representations of language unity and introduces \emph{language diversity} as formal data, both in terms of lexical model and actual content.

The UKC data model, the theoretical underpinnings of which have been exposed in \cite{giunchiglia2018one}, is illustrated in Figure~\ref{fig:ukc_structure}. On the top of the figure, a supra-lingual \emph{concept layer} contains hierarchies of concepts that represent lexical meaning shared across languages. Concepts thus act as bridges across languages. The only criterion for a concept to be present in the concept layer is that it is lexicalised by at least one language.

The bottom \emph{lexicon layer} consists of language-specific lexicons. As in other lexical databases, these provide lexicalisations for concepts, such as the English \emph{`rice'} and the Italian \emph{`riso'} for the concept of \emph{rice} in Figure~\ref{fig:ukc_structure}. Beyond lexicalisations, however, the UKC lexicons also provide rich cross-lingual information on language unity and diversity through the following constructs.

\para{Lexical gaps.} As mentioned in the introduction, if not addressed properly, lexical untranslatability negatively affects the performance of cross-lingual applications. Erroneous Google translations, such as the Hungarian sentence
\begin{center}
\emph{`A bátyám három évvel fiatalabb nálam,'}
\end{center}
meaning `my older brother is three years younger than me,' are systematically produced due to  non-existent equivalent translations, in this case for \emph{brother} Hungarian.\footnote{Apart from the laborious and thus rarely used \emph{fiútestvér}.} 
Likewise, as shown in Figure~\ref{fig:ukc_structure}, English has no single word for \emph{raw, uncooked rice} while Swahili has no word equivalent to the general term \emph{rice}. The UKC provides evidence of untranslatability by representing lexical gaps inside lexicons. Such information can be used, among others, to indicate the absence of equivalent terms to downstream cross-lingual applications.

\para{Cross-lingual sense relations.} Beyond providing shared word meanings as other MLDBs do, the UKC represents a richer set of interlingual connections between word senses. For example, in Figure~\ref{fig:ukc_structure}, the English \emph{`rice'} and the Italian \emph{`riso'} are connected through a \emph{cognate} relationship.\footnote{Cognates are words in different languages that sound the same and have the same (or similar) meaning due to a common etymological origin.} Such information can be exploited as evidence of cross-lingual similarity, e.g.~as seeds in cross-lingual tasks such as bilingual lexicon induction \cite{batsuren2021large}.

\para{Metadata on language diversity.} Beyond standard typological metadata such as language phylogeny or the geographical locations of speakers, the UKC also integrates cross-linguistic metadata computed from its own lexico-semantic content. Based on cross-lingual cognate relationships, we computed large-scale \emph{lexicon similarity} data across 27~thousand language pairs over 331~languages. Lexicon similarity \cite{bella2021database} formally characterises the extent to which the vocabularies of two languages `resemble each other', taking differing writing systems and orthographies into account. This metric has, in our view, a better potential in predicting the success of cross-lingual tasks (such as transfer learning or joint supervised training) than language phylogeny, as it is based on the overlaps of contemporary lexicons as opposed to historical relatedness.

\para{Language-specific word meanings.} Diversity also means acknowledging our partial understanding of how specific languages conceptualise lexical meaning and the ultimate impossibility of an exhaustive interlingual model. The UKC is the only lexical database to allow the co-existence of shared and language-specific word meaning hierarchies, inside the concept layer and the lexicons, respectively. Figure~\ref{fig:ukc_structure} shows culture-specific words and meanings (of rice-based foods) represented inside the Kannada lexicon, not yet integrated into the shared concept layer.

\para{Language-specific lexical relations.} Lexical relations within individual languages are sometimes part of lexical databases, such as antonymy or derivation in the Princeton WordNet \cite{miller1998wordnet}. The UKC introduces relation types not typically part of lexical databases---such as \emph{metonym-of} or \emph{homograph-of}---and provides corresponding relation instances in multiple languages.

\begin{figure*}
\begin{center}
\includegraphics[width=\textwidth]{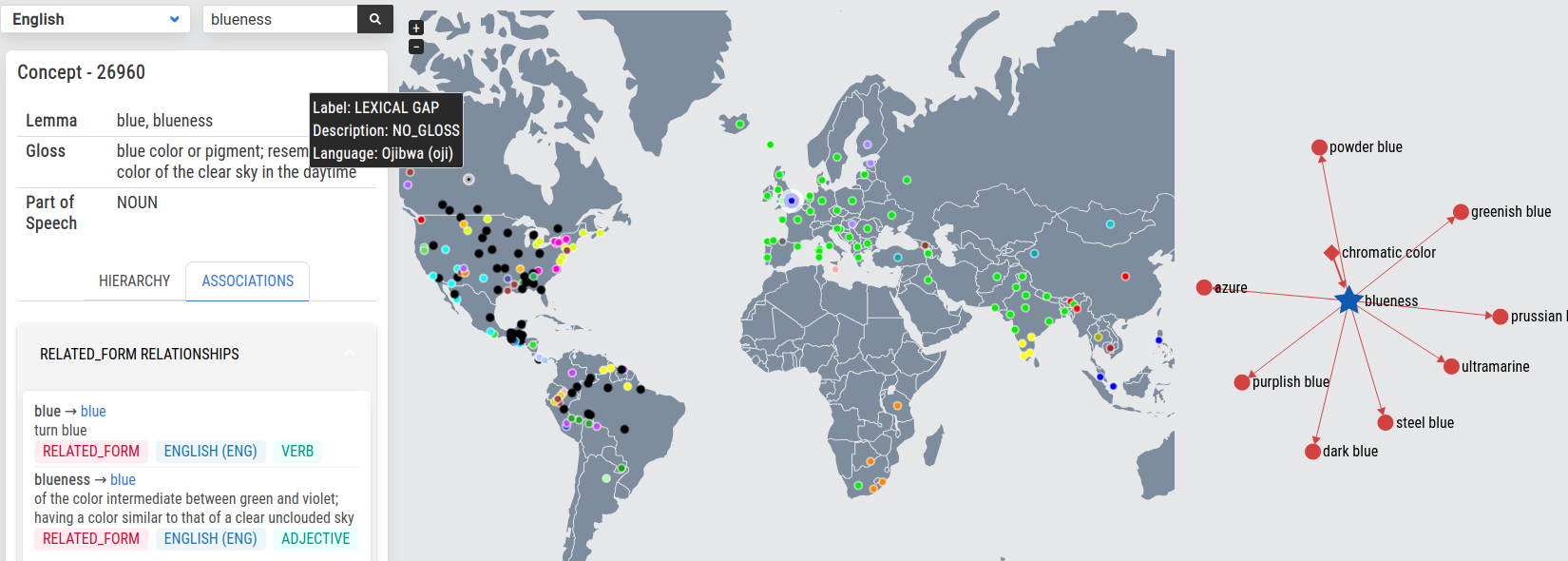}
\caption{\label{fig:concept_browse}Exploring the concept of \emph{blueness} as lexicalised in the English language (left), in the world (middle), and as part of the supra-lingual concept hierarchy (right).}
\end{center}
\end{figure*}

Table~\ref{tab:ukc_content} shows the current contents of the UKC (as of January~2022) in terms of the data types enumerated above. Concepts and concept relations were initially derived from the Princeton WordNet, as in all other MLDBS, but then extended with 400~new concepts and 490~new relations (research awaiting to be published). We obtained lexicalisations from Wiktionary, NorthEuraLex \cite{dellert2020northeuralex}, the \emph{Native Languages of the Americas}\footnote{\url{http://www.native-languages.org}}, as well as the wordnet resources cited at the end of our paper. \nocite{bella2020major,batsuren2019building,ganbold2018using,bhattacharyya2010indowordnet,griesel2014taking,regragui2016arabic,noor2011creating,osenova2018data,wang2013building,huang2010chinese,raffaelli2008building,wu2021dannet,linden2010finnwordnet,grigoriadou2004software,isahara2008development,garabik2013multilingual,gonzalez2012multilingual,fjeld2009nornet,postma2016open,rademaker2016introducing,keyvan2007developing,maziarz2012approaching,tufics2007romanian,garabik2010slovak,fivser2012slownet,borin2008hunting,thoongsup2009thai,sagot2008construction,bentivogli2004revising} Lexical gaps were obtained mostly through our original research (about~18k relating to diversity-rich domains such as kinship and colours, not yet published, but also 600~gaps from \cite{bella2020major}), and in a minor part from the few third-party resources providing such information \cite{pianta2002multiwordnet,ordan2007hebrew}. We computed cross-lingual sense relations from UKC data, reusing our method published in \cite{batsuren2019cognet,batsuren2021large}. Language-specific sense relations were obtained in a minor part from wordnets providing such data (48k~relations), and in a major part from our own research on multilingual morphology  \cite{batsuren2021morphynet} (770k~derivations in 16~languages) and metonymy (not yet published, 25k~metonyms in 191~languages). Finally, lexicon-level metadata on language diversity combines online sources \cite{wals} with results of our own research on the similarity of lexicons \cite{bella2021database}.

\begin{table}
\setlength{\tabcolsep}{3pt}
\begin{tabular}{p{5.6cm}r}
\hline
\textbf{Content type} & \textbf{Data size}  \\\hline
Languages       & 2,176 \\
Concepts        & 106k \\
Concept relations & 109k \\
Lexicalisations (word senses) & 2.8M \\
Lexical gaps    & 24k \\
Cross-lingual sense relations & 8M \\
Language-specific relations & 840k \\
Lexicon-level diversity metadata & 30k \\\hline
\end{tabular}
\caption{\label{tab:ukc_content}UKC contents by type and their provenance.}
\end{table}

\section{Exploring Diversity Data}

The website of the UKC database provides browseable online access to the full database contents, data visualisation tools, extensive information on related projects, publications, source materials, as well as example downstream services, such as word translation between any two languages or multilingual word sense disambiguation (to be released soon).

A major feature of the website is the interactive exploration of lexicons and diversity data. The user can browse: (1)~linguistic metadata of the 1.2k~lexicons, selecting the language from an interactive map (Figure~\ref{fig:ukc_map}) or by name; (2)~within a language, all meanings of a word typed in by the user; and (3)~lexicalisations and gaps of a concept in the current language and in all languages of the world.

A screenshot of the last---and richest---concept exploration functionality, taking the example concept of \emph{blueness}, is provided in Figure~\ref{fig:concept_browse}. On the left-hand side of the screen, details are provided on the lexicalisation of the concept in the current language, such as synonyms, definition, part of speech, as well as lexical relationships to other word senses (e.g.~derivations, metonyms, cognates in other languages). The middle part of the screen shows an interactive clickable map of all languages that either lexicalise the concept in the database or, on the contrary, \emph{are known not to lexicalise it}. The colour-coded dots (indicating the language family while black stands for a gap) thus provide an instant global typological overview for the concept selected, e.g.~from Figure~\ref{fig:concept_browse} one can see that a lot of languages in the Americas do not lexicalise blue as a separate colour. Some language do not appear on the map due to lexicon incompleteness: for those languages the UKC has no information whether they lexicalise \emph{blueness} or not.

The right-hand side, finally, shows the concept in the context of the concept hierarchy, shown as an interactive graph: the currently observed concept \emph{blueness}, as well as the parent (broader) and child (narrower) concepts. Other lexico-semantic relationships (e.g.~meronymy and metonymically related concepts) are also shown when they exist. While for usability reasons the graph only displays a part of the full hierarchy, it is navigable, allowing the entire concept graph to be explored in the currently selected language. Changing the language is as simple as clicking on the map or selecting it from the drop-down in the upper left corner of the screen. Colours in the graph are indicative of language diversity: they show whether a concept is lexicalised in the current language (dark-coloured nodes), are missing from its lexicon (light-coloured nodes), or are lexical gaps (black nodes).

\section{Visualising Language Diversity}

Beyond the fine-grained word and concept exploration presented in the previous section, the UKC website also offers visualisation tools that allow humans to grasp diversity both in its globality and from different angles. Currently the following tools are provided, three of which we present below: (1)~cognate diversity clusters; (2)~colexifications; (3)~a gap explorer for a fixed set of domains that are lexically diverse; (4)~lexical similarity graphs; and (5)~visual statistics.

\begin{figure*}[h!]
    \centering
    \includegraphics[width=\textwidth]{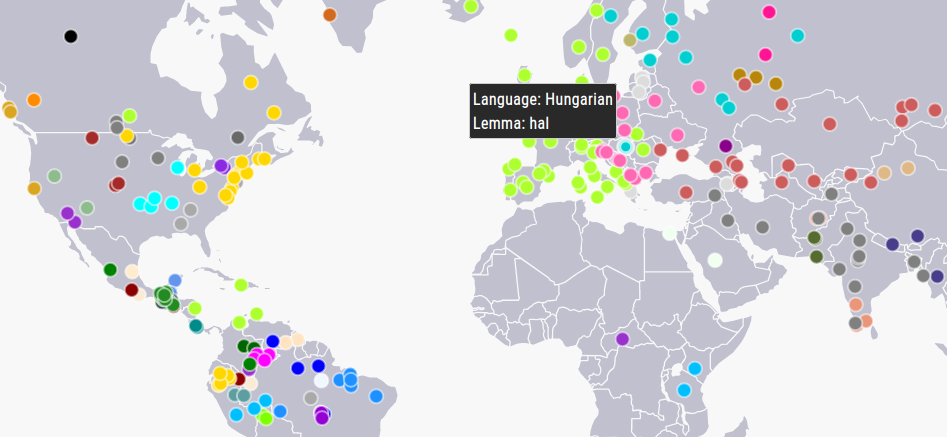}
    \caption{Cognate clusters for the concept of \emph{fish}.}
    \label{fig:cognate_map}
\end{figure*}

\para{Cognate diversity clusters.} This tool shows \emph{cognate clusters} on the map for a given concept selected by the user, computed from cognate data inside the UKC. In Figure~\ref{fig:cognate_map}, the concept of \emph{fish} is selected: each dot represents a lexicon that contains a word for \emph{fish}. Two dots are of the same colour if the two words are cognates of each other. For example, the English \emph{`fish'} and the Italian \emph{`pesce'} are within one cognate cluster (in light green in the figure) while the Hungarian \emph{`hal'} and the Finnish \emph{`kala'} are in another cluster (in turquoise). The number and distribution of clusters for a given concept provide information about its universality or diversity: \emph{coffee} is a so-called \emph{universal concept} while \emph{woman} is an extremely diverse one.

\begin{figure*}[h!]
    \centering
    \includegraphics[width=\textwidth]{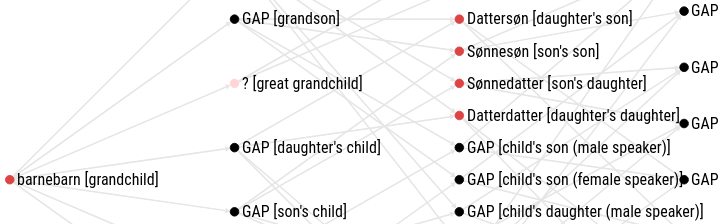}
    \caption{Detail of the \emph{grandchild} subdomain as lexicalised by the Danish language.}
    \label{fig:gap_graph}
\end{figure*}

\para{Lexical gap explorer.} Certain domains---such as kinship, food, colours, or body parts---are known by linguists to be lexically diverse, for reasons related to culture, geography, but also grammar and other factors \cite{lehrer1970notes}. The \emph{gap explorer tool} displays a full concept hierarchy for a domain or subdomain selected by the user. Figure~\ref{fig:gap_graph} shows the UKC concept structure of the subdomain of \emph{siblings} from the \emph{kinship} domain. For the language selected (Danish in the figure), the tool displays existing lexicalisations, indicates incompleteness (missing word), and provides known lexical gaps. This allows for quick comparisons of how different languages lexicalise (or not) a given domain. For example, for the \emph{cousins} subdomain that consists of 67~concepts, English only lexicalises the root concept \emph{cousin} with all other concepts as gaps, while South Indian languages provide no less than 16~distinct words depending on the age, sex, and lineage (patrilineal/matrilineal) of the cousin.

\begin{figure*}[h!]
    \centering
    \includegraphics[width=\textwidth]{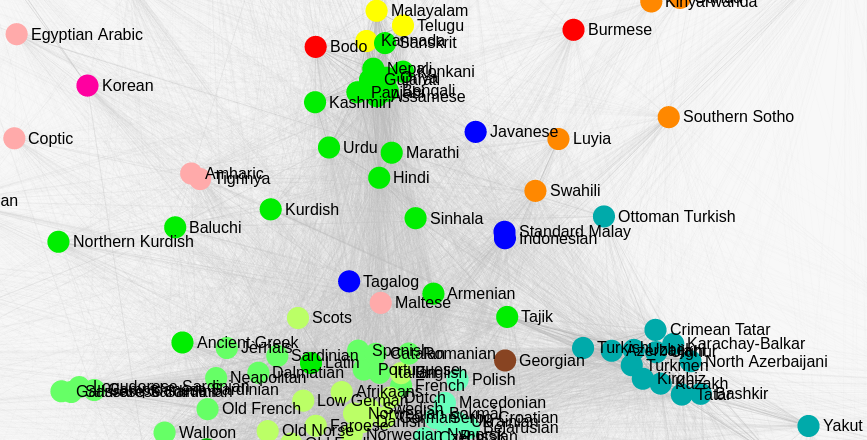}
    \caption{Detail from the lexical similarity graph coloured according to language families.}
    \label{fig:similarity_graph}
\end{figure*}

\para{Lexical similarity graphs.} Relying on the extensive lexical data inside the UKC, we compute pairwise similarities between languages based on the amount of shared cognates, using the method described in our recent paper \cite{bella2021database}. In order to interpret the resulting similarity data for humans, i.e.~provide a global overview of lexical similarity, we compute a dynamic graph visualisation where nodes are languages and edge lengths are proportional to lexical similarities. The graph computation relies on a physical model of attraction and repulsion among nodes, using the \emph{ForceAtlas2} library \cite{jacomy2014forceatlas2}. We provide two distinct colourings for the same graph: one based on language families (shown in Figure~\ref{fig:similarity_graph}) and the other based on geographical distance. These graphs visualise how the similarity of contemporary lexicons correlates with (historic) phylogeny and with the geographical closeness of speakers. As a way to make language evolution visual, we also provide the equivalent graph computed over data from historical linguistics, obtained from the ASJP database \cite{asjp}. Insights gained from these graphs may also help computational linguists predict the performance of automated tasks that involve some form of lexicon mapping (e.g.~bilingual lexicon induction or machine translation) over specific language pairs.

\begin{figure*}[t]
    \centering
    \includegraphics[width=.8\textwidth]{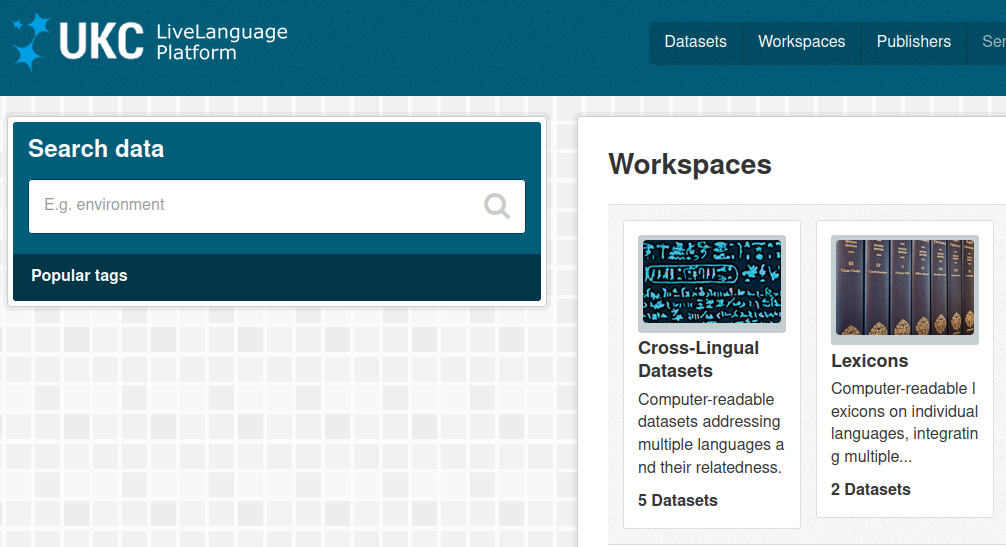}
    \caption{The UKC LiveLanguage Data Catalogue.}
    \label{fig:livelanguage}
    \vspace{-2mm}
\end{figure*}

\section{The LiveLanguage Data Catalogue}


As a complement to online exploration, we are also making available the contents of the UKC for computational applications. While an open, fine-grained, API-based access to the data is planned as future work, we are already in the process of publishing data for download through the \emph{UKC LiveLanguage} data catalogue.\footnote{\url{http://www.livelanguage.eu}} The catalogue, accessible from the website through any of the numerous download links, provides access to the UKC data through multiple modalities. Accordingly, the structure of the catalogue, shown in Figure~\ref{fig:livelanguage}, consists of (1)~\emph{cross-lingual datasets} from projects related to the UKC; (2)~\emph{individual lexicons} that incorporate all types of data (as in Section~2) that describe a single language; (3)~\emph{lexicon sets} consisting of multiple lexicons as well as of cross-lingual relationships among them. All datasets are published in full respect of the licensing constraints of their constituting resources; data that disallow redistribution are excluded from the catalogue.

\para{Raw data on cross-lingual diversity.} These datasets, produced by the authors as ongoing or past projects on diversity, cover domain-specific lexical gaps, multilingual morphology, lexical similarity, and cognate relationships. Datasets are distributed in their original (e.g.~tab-separated) formats, with concepts linked to Princeton WordNet~3.0 identifiers for interoperability with third-party data.  

\para{Individual lexicons.} These datasets are produced as language-specific `cross-sections' of the full UKC data. Their added value lies in the integration of multiple sources---words from wordnets and Wiktionary, language-specific morphological and lexico-semantic relationships, gaps---into a single formal representation. These datasets will be provided in multiple formats (under development), including the ISO standard Lexical Markup Framework (LMF) format as well as OntoLex.

\para{Lexicon sets.} The notion of language diversity gains full significance \emph{across} languages. Consequently, the development of an online service is underway to allow the simultaneous download of multiple concept-aligned lexicons as a single multilingual resource. The service will export multilingual data from the UKC database in real time. Such datasets will be directly exploitable in cross-lingual training and evaluation tasks.

\section{Conclusions and Future Work}

With the UKC database, website, and data catalogue, we hope to contribute to the exploitation of language diversity as computational data. All components of the system are going through a rapid evolution: the database contents in terms of language support, lexicon correctness and completeness, the data exploration and visualisation tools, a new set of demonstrators, APIs, and downloadable datasets are continually being created and extended. At the same time, the exploitation of diversity data to improve state-of-the-art cross-lingual applications is a research direction that we expect to gain importance in the near future.

\bibliography{custom}
\bibliographystyle{acl_natbib}




\end{document}